\documentclass[journal,comsoc]{IEEEtran}
\usepackage[T1]{fontenc}
\usepackage{commath}

\usepackage{times}
\usepackage{soul}
\usepackage{url}
\usepackage[colorlinks,linkcolor={blue},pagebackref=true,breaklinks=true,colorlinks,bookmarks=false]{hyperref}
\usepackage[utf8]{inputenc}
\usepackage[small]{caption}
\usepackage{graphicx}
\usepackage{amsmath}
\usepackage{amssymb}
\usepackage{xspace}
\usepackage{booktabs}
\usepackage{commath}
\newcommand{\ie}{\textit{i}.\textit{e}.,}

\newcommand{\etal}{\textit{et al}. }

\def\ourmodel{MSGCN}
\def\ange{AGE}

\newcommand{\STAB}[1]{\begin{tabular}{@{}c@{}}#1\end{tabular}}

\usepackage[font=bf,labelfont=bf]{caption}
\usepackage{subcaption}
\usepackage{makecell}
\usepackage{multirow}
\usepackage{booktabs}

\usepackage[table, svgnames]{xcolor}
\definecolor{Gray}{gray}{0.85}

\begin{document}

\title{
Fusing Higher-order Features in Graph Neural Networks for 
Skeleton-based Action Recognition
}

\author{Zhenyue~Qin,
        Yang~Liu,
        Pan~Ji, 
        Dongwoo~Kim, 
        Lei Wang, 
        R.I. (Bob) McKay, 
        Saeed~Anwar,
        Tom~Gedeon
\thanks{Z. QIN and R. McKay were with Australian National University (ANU). Y. LIU, L. Wang, and S. ANWAR were with both ANU and Data61, CSIRO. P. JI was with Tencent XR Lab. D. KIM was with POSTECH. T. GEDEON was with Optus-Curtin Centre of Excellence in AI, Curtin University. Corresponding authors: Yang Liu and Zhenyue Qin. Emails: yang.liu3@anu.edu.au, zhenyue.qin@anu.edu.au.}% 
}

% The paper headers
% \markboth{Special Issue on Deep Neural Networks for Graphs: Theory, Models, Algorithms and Applications}%
\markboth{}%
{Shell \MakeLowercase{\textit{et al.}}: Bare Demo of IEEEtran.cls for IEEE Communications Society Journals}

% make the title area
\maketitle

% As a general rule, do not put math, special symbols or citations
% in the abstract or keywords.
\begin{abstract}
Skeleton sequences are lightweight and compact, and thus are ideal candidates for action recognition on edge devices. Recent skeleton-based action recognition methods extract features from 3D joint coordinates as spatial-temporal cues, using these representations in a graph neural network for feature fusion to boost recognition performance. The use of first- and second-order features, \ie{} joint and bone representations, has led to high accuracy. Nonetheless, many models are still confused by actions that have similar motion trajectories. To address these issues, we propose fusing higher-order features in the form of angular encoding into modern architectures to robustly capture the relationships between joints and body parts. This simple fusion with popular spatial-temporal graph neural networks achieves new state-of-the-art accuracy in two large benchmarks, including NTU60 and NTU120, while employing fewer parameters and reduced run time. Our source code is publicly available at: 
\href{https://github.com/ZhenyueQin/Angular-Skeleton-Encoding}{https://github.com/ZhenyueQin/Angular-Skeleton-Encoding}.

\end{abstract}

% Note that keywords are not normally used for peerreview papers.
% \begin{IEEEkeywords}
% Human action recognition, graph neural network, video understanding
% \end{IEEEkeywords}

\IEEEpeerreviewmaketitle

\section{Introduction}
\label{sec:Intro}

Skeleton-based action recognition is more robust to background information and easier to process, attracting increasing attention~\cite{shi2019skeleton} in the community. Recently, deep graph neural networks fuel the recent surge of accuracy for skeleton-based action recognition~\cite{yan2018spatial}. By leveraging graph neural networks, action recognizers more thoroughly extract the topological information within the skeleton sequences. 

To make graph neural networks applicable for skeleton-based action recognition, skeletons are treated as graphs, with each vertex representing a body joint and each edge a bone. Initially, only first-order features were employed, representing the coordinates of the joints~\cite{yan2018spatial}. Subsequently, \cite{shi2019two} introduced a second-order feature: each bone is expressed as the vector difference between one joint's coordinate and that of its nearest neighbor in the direction of the body center. Their experiments show that these second-order features improve the recognition accuracy of skeleton-based action recognizers.

\begin{figure}[ht]
\centering
\begin{subfigure}{.21\textwidth}
  \centering
  % include first image
  \includegraphics[width=.99\linewidth]{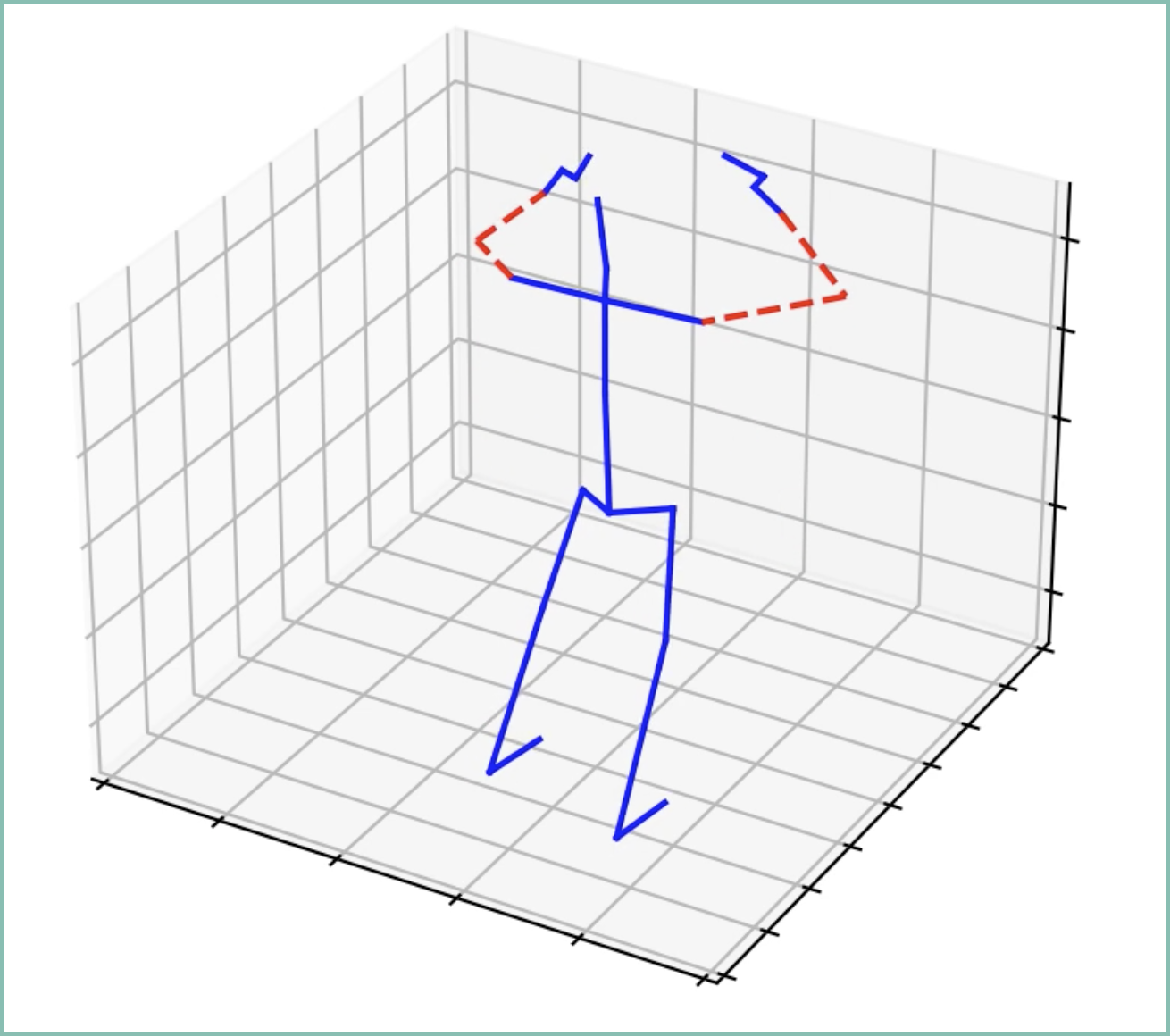}  
  \caption*{Taking off glasses }
  \label{fig:off_headphone}
\end{subfigure}
\begin{subfigure}{.21\textwidth}
  \centering
  % include second image
  \includegraphics[width=.99\linewidth]{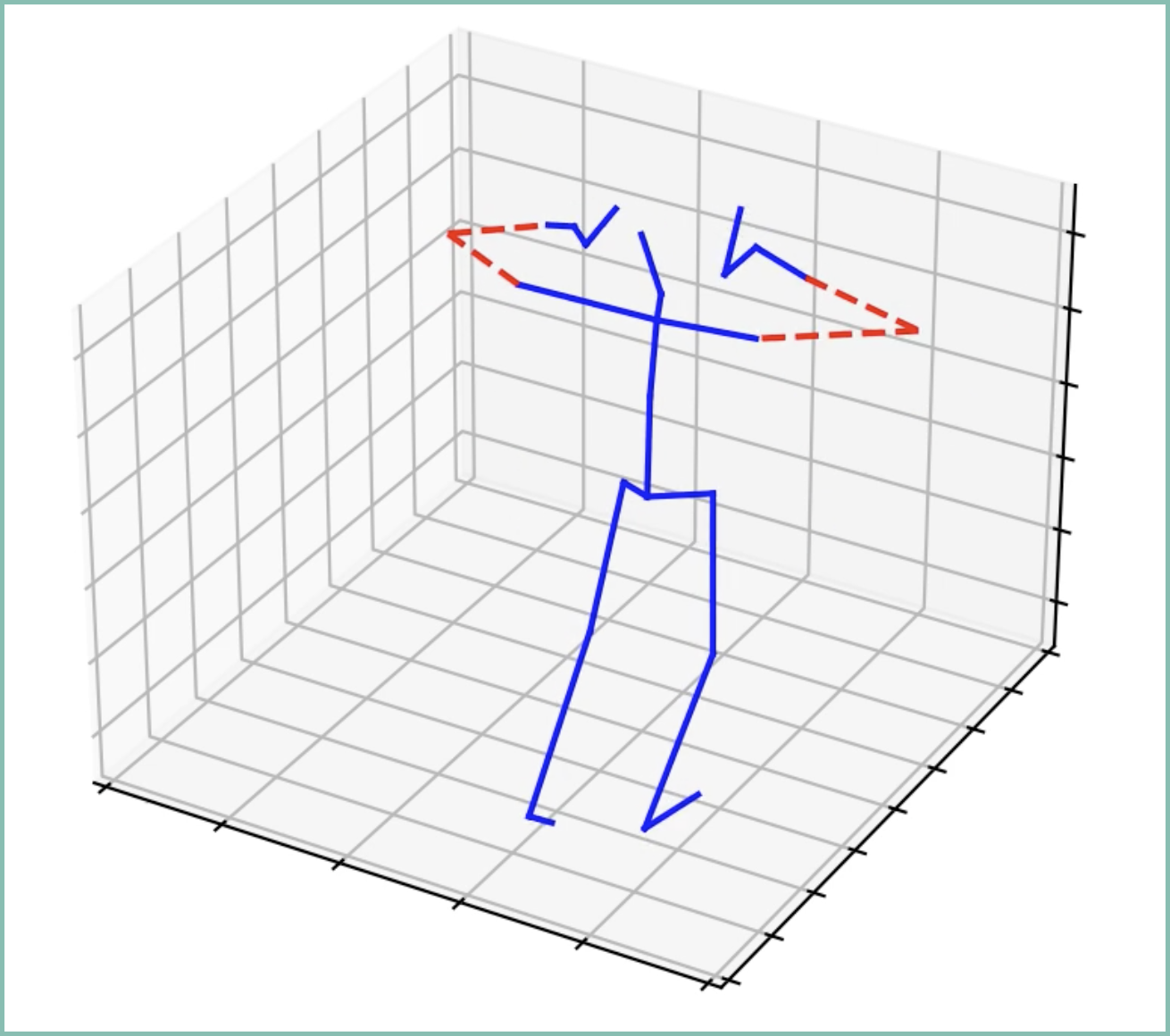}  
  \caption*{Taking off headphones }
  \label{fig:off_glass}
\end{subfigure}
\caption{Sample skeletons with similar motion trajectories: (left) taking off glasses vs (right) taking off headphones. 
The angles formed by red dashed lines (\ie{} the fore- and upper arms) are distinctive, which are informative in distinguishing these two similar motions.
}
% \vspace{-4mm}
\label{fig:skeleton_off_glass_headphone}
\end{figure}

However, existing methods suffer from the poor performance of discriminating actions with similar motion trajectories (see ~\autoref{fig:skeleton_off_glass_headphone}).
Since the joint coordinates in each frame are similar in these actions, it is challenging to identify the cause of nuances between coordinates. It can be due to various body sizes, motion speeds, or actually performing different actions. 
To robustly capture the relative movements between body parts while maintaining invariance for different body sizes of human subjects, in this paper, we propose the use of higher-order representations in the form of angles. We refer to the new proposed feature as angular encoding, which can be applied to both static and velocity domains of human body joints. 
Thus, the proposed encoding allows the model to recognize actions more precisely.
Experimental results reveal that by fusing angular information into the existing modern action recognition architectures, such as Spatio-Temporal Graph Convolutional Network (STGCN)~\cite{yan2018spatial} and Decoupling GCN~\cite{chengdecoupling}, confusing action sequences can be classified more accurately, especially when the actions have very similar motion trajectories. 

It is worth considering whether it is possible to design a neural network to implicitly learn angular features. 
However, such a design would be challenging for current graph convolutional networks (GCNs)~\cite{wu2020comprehensive,velivckovic2017graph}, mainly due to two reasons. (a) \textit{Conflicts between more layers and higher performance of GCNs}: GCNs are currently the best-performing models in classifying skeleton-based actions. To model the relationships among all the joints, a graph network requires many layers. However, recent work implies the performance of a GCN can be compromised when it goes deeper
due to over-smoothing problems~\cite{min2020scattering}. (b)
\textit{
Limitation of adjacency matrices}: recent graph networks for action recognition learn the relationships among nodes via an adjacency matrix, which only captures pairwise relevance, whereas angles are third-order relationships involving three related joints.

We summarize our contributions as follows: 
\begin{enumerate}
    \item We propose a rich collection of higher-order representations in the form of the angular encoding defined in both static and velocity domains. The encoding captures relative motion between body parts while maintaining invariance against different human body sizes. 
    \item The angular features can be easily fused into existing action recognition architectures to further boost performance. Our experiments show that angular features are complementary information relative to existing features, \ie{} the joint and bone representations. 
    \item We are the first to incorporate multiple categories of angular features into modern spatial-temporal GCNs and achieve state-of-the-art results on several benchmarks, including NTU60 and NTU120. Meanwhile, if a simple model (employing fewer training parameters and requiring less inference time) has equipped with the proposed angular encoding, it becomes powerful. Thus, the proposed angular encoding supports real-time action recognition on edge devices. 
\end{enumerate}

\begin{figure*}
    \centering
    \includegraphics[width=0.99\linewidth]{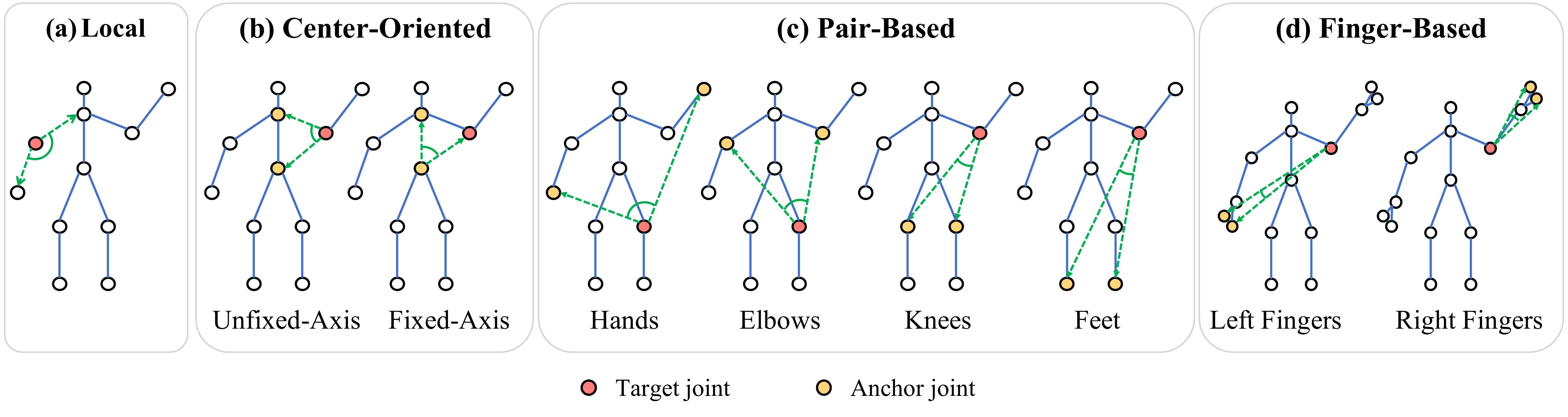}
    \caption{The proposed four types of angular features. We extract angular features for the target joint (in red dots) which corresponds to the root of an angle. The anchor joints (in yellow dots) are fixed endpoints of angles. Green dashed lines represent the two sides of an angle.}
    \label{fig:skeletons}
\end{figure*}
\begin{figure*}[ht]
    \centering
    \includegraphics[width=.99\linewidth]{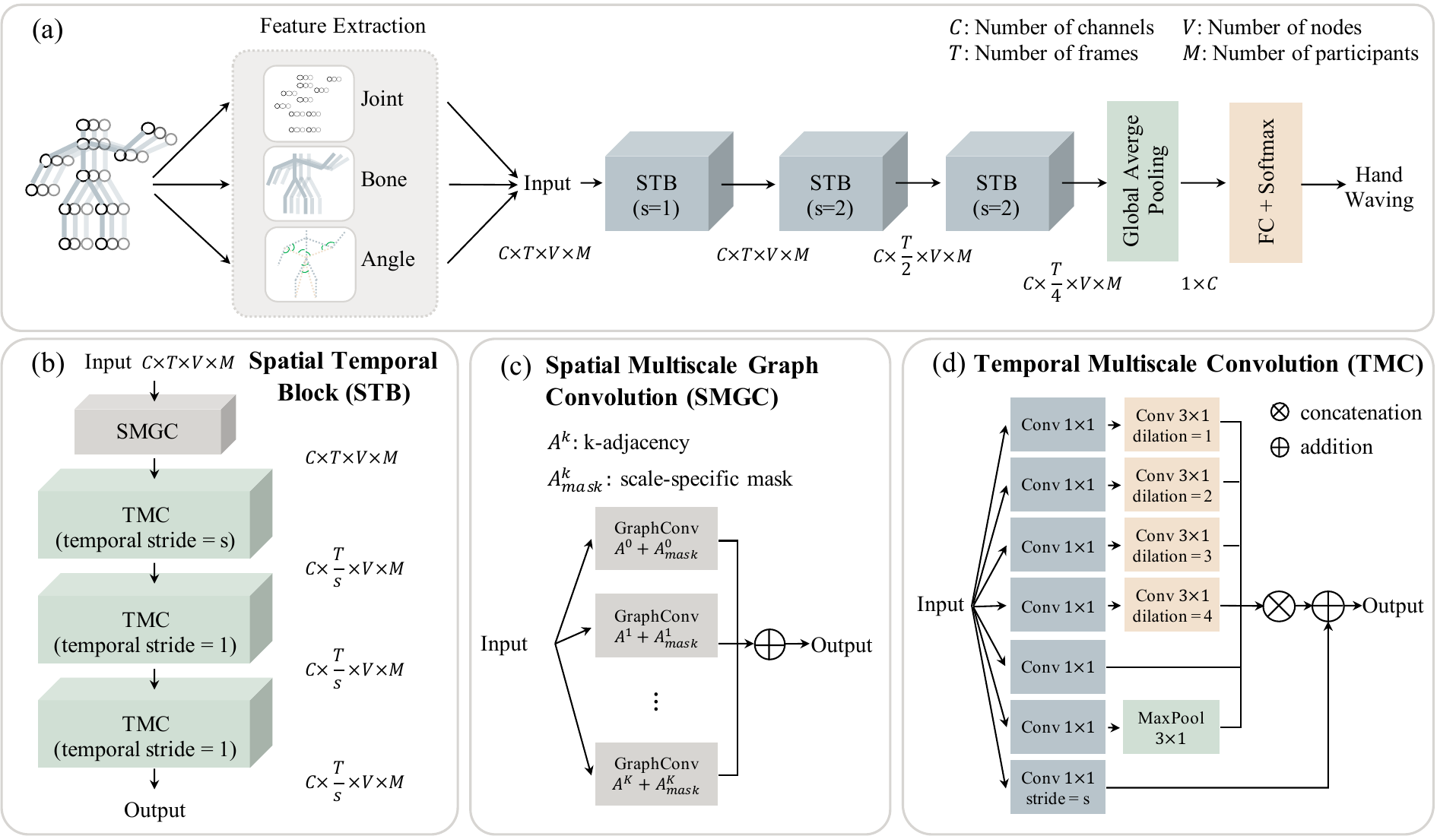}
    \caption{
    Our backbone architecture is composed of three spatial-temporal blocks, each consisting of a spatial multiscale graph convolution and a temporal multiscale convolution unit. The spatial multiscale unit extracts structural skeleton information with parallel graph convolutional layers. The temporal multiscale unit draws correlations with four functional groups. See Section~\ref{sec: angnet} for more details. 
    }
    \label{fig:net_architecture}
\end{figure*} 

\begin{table*}[t]
\caption{
Comparison of recognition performance on four settings of two benchmark datasets. We compare not only the recognition accuracy but also the total number of parameters (\#params) in the networks. 
\#Ens is the number of models used in an ensemble. 
BSL means to use the original feature without employing angular encoding.
\ange{}-S and \ange{}-V stand for concatenating the original representation with angular encoding in the static and velocity domains respectively. 
Joint/J and Bone/B denote the use of joint and bone features respectively.
The top accuracy is highlighted in red bold, and the second best performance is highlighted in blue.
Symbol \& indicates ensembling models trained with different input features given in the parenthesis. GFlops stands for the floating-point operations performed by a model, which is the number of multiply-add operations that a model performs. 
} 
\centering

\resizebox{0.99\textwidth}{!}{
\begin{tabular}{l c c | rrrr | rrrr | r r}
\toprule
% row 1
%\rowcolor{Gray}
&
&
&
\multicolumn{4}{c| }{\textbf{NTU60}} 
& \multicolumn{4}{c|}{\textbf{NTU120}}
& \multicolumn{1}{c }{\textbf{\# Params}}
& 
\\
% row 2
\cline{4-11}
\multirow{-2}{*}{\textbf{Methods}} &
\multirow{-2}{*}{\textbf{Year}} &
\multirow{-2}{*}{\textbf{\# Ens}}  &
X-Sub
&  Acc $\uparrow$
& X-View
&  Acc$\uparrow$ 
& X-Sub
&  Acc$\uparrow$ 
& X-Set
&  Acc$\uparrow$
& \multicolumn{1}{c }{\textbf{(M)}}
& \multirow{-2}{*}{\textbf{GFlops}}
\\
\midrule 
\rowcolor{Gray!30}
HCN~\cite{2018_ijcai_hcn} & 2018 & 1 & 86.5 & - & 91.1 & - & - & - & - & - & - & -
\\ 
MAN~\cite{xie2018memory} & 2018 & 1 & 82.7 & - & 93.2 & - & - & - & - & - & - & -
\\ 
\rowcolor{Gray!30}
ST-GCN~\cite{yan2018spatial} & 2018 & 1 & 81.5 & - & 88.3 & - & - & - & - & - & 2.91 & 16.4
\\ 
AS-GCN~\cite{li2019actional} & 2019 & 1 & 86.8 & - & 94.2 & - & - & - & - & - & 7.17 & 35.5
\\
\rowcolor{Gray!30}
AGC-LSTM~\cite{si2019attention} & 2019 & 2 & 89.2 & - & 95.0 & - & - & - & - & - & - & -
\\
2s-AGCN~\cite{shi2019two} & 2019 & 4 & 88.5 & - & 95.1 & - & - & - & - & - & 6.72 & 37.2
\\
\rowcolor{Gray!30}
DGNN~\cite{shi2019skeleton} & 2019 & 4 & 89.9 & - & 96.1 & - & - & - & - & - & 8.06 & 71.1
\\
Bayes-GCN~\cite{zhao2019bayesian} & 2019 & 1 & 81.8 & - & 92.4 & - & - & - & - & - & - & -
\\
\rowcolor{Gray!30}
SGN~\cite{zhang2020semantics} & 2020 & 1 & 89.0 & - & 94.5 & - & 79.2 & - & 81.5 & - & 0.69 & 15.4
\\
% Shift-GCN~\cite{cheng2020skeleton} & 2020 & 4 & 90.7 & \color{blue}{96.5} & 85.9 & 87.6 & 2.77 & 19.2 //
DeCoupleGCN~\cite{chengdecoupling} & 2020 & 4 & 90.8 & - & \color{red}{\textbf{96.6}} & - & 86.5 & - & 88.1 & - & 13.72 & 102.3
\\
\rowcolor{Gray!30}
MS-G3D~\cite{liu2020disentangling} & 2020 & 2 & 91.5 & - & 96.2 & - & 86.9 & - & 88.4 & - & 6.44 & 98.0
\\
MST~\cite{chen2021multi} & 2021 & 2 & 91.1 & - & 96.4 & - & 87.0 & - & 88.3 & - & - & -
\\
\rowcolor{Gray!30}
AdaSGN~\cite{shi2021adasgn} & 2021 & 4 & 90.5 & - & 95.3 & - & 85.9 & - & 86.8 & - & - & -
\\
Ta-CNN~\cite{xu2022topology} & 2022 & 2 & 90.7 & - & 95.1 & - & 85.7 & - & 87.3 & - & - & -
\\
\rowcolor{Gray!30}
Efficient-Self-Attention~\cite{qin2022efficient} & 2022 & 4 & 90.5 & - & 96.1 & - & 85.7 & - & 86.8 & - & - & -
\\
\midrule 
\multicolumn{13}{c}{Our Methods} \\
% Static Joint 
\midrule 
\rowcolor{Gray!30}
BSL-S (Joint) & - & 1 & 87.2 & - & 93.7 & - & 81.9 & - & 83.5 & - & 1.42 & 19.0 \\
\ange{}-S (Joint) & - & 1 & 88.7 & 1.5 & 94.5 & 0.8 & 83.2 & 1.3 & 83.7 & 0.2 & 1.44 & 19.4 \\
% Static Bone
\midrule 
\rowcolor{Gray!30}
BSL-S (Bone) & - & 1 & 88.2 & - & 93.6 & - & 84.0 & - & 85.3 & - & 1.42 & 19.0 \\
\ange{}-S (Bone) & - & 1 & 89.2 & 1.0 & 94.8 & 1.2 & 84.6 & 0.6 & 85.5 & 0.2 & 1.44 & 19.4 \\
% Velocity Joint 
\midrule 
\rowcolor{Gray!30}
BSL-V (Joint) & - & 1 & 86.0 & - & 93.3 & - & 79.3 & - & 80.8 & - & 1.42 & 19.0 \\
\ange{}-V (Joint) & - & 1 & 88.2 & 2.2 & 94.5 & 1.2 & 81.8 & 2.5 & 83.7 & 2.7 & 1.44 & 19.4 \\
% Velocity Bone
\midrule
\rowcolor{Gray!30}
BSL-V (Bone) & - & 1 & 86.4 & - & 92.7 & - & 80.3 & - & 82.0 & - & 1.42 & 19.0 \\
\ange{}-V (Bone) & - & 1 & 88.0 & 1.6 & 94.8 & 2.1 & 82.9 & 2.6 & 85.1 & 3.1 & 1.44 & 19.4 \\
% Static Joint + Bone
\midrule
\rowcolor{Gray!30}
BSL-S (Joint+Bone) & - & 1 & 89.2 & - & 95.1 & - & 84.1 & - & 86.0  & - & 1.44 & 19.4 \\
\ange{}-S (Joint+Bone) & - & 1 & 90.0 & 0.8 & 95.2 & 0.1 & 85.9 & 1.8 & 86.8 & 0.8 & 1.46 & 19.6 \\
% Velocity Joint + Bone
\midrule
\rowcolor{Gray!30}
BSL-V (Joint+Bone) & - & 1 & 86.1 & - & 92.6 & - & 80.5 & - & 81.5 & - & 1.44 & 19.4 \\
\ange{}-V (Joint+Bone) & - & 1 & 87.1 & 1.0 & 94.0 & 1.4 & 83.0 & 2.5 & 84.6 & 3.1 & 1.46 & 19.6 \\
\midrule 
% Ensemble
\rowcolor{Gray!30}
BSL-Ens: S(J)\&V(J) & - & 2 & 89.3 & - & 94.7 & - & 84.3 & - & 85.2 & - & 2.84 & 38.0 \\ 
\ange{}-Ens: S(J)\&V(J) & - & 2 & 90.5 & 1.2 & 95.5 & 0.8 & 85.3 & 1.0 & 85.8 & 0.6 & 2.88 & 38.8 \\ 
\midrule 
\rowcolor{Gray!30}
BSL-Ens: S(B)\&V(B) & - & 2 & 90.5 & - & 94.7 & - & 86.3 & - &  85.6 & - & 2.84 & 38.0 \\ 
\ange{}-Ens: S(B)\&V(B) & - & 2 & 90.8 & 0.3 & 95.5 & 0.8 & 87.3 & 1.0 & 86.8 & 1.2 & 2.88 & 38.8 \\ 
\midrule 
\rowcolor{Gray!30}
BSL-Ens: S(J+B)\&V(J+B) & - & 2 & 90.5 & - & 95.7 & - & 86.4 & - & 86.4 & - & 2.88 & 38.8 \\ 
\ange{}-Ens: S(J+B)\&V(J+B) & - & 2 & 91.0 & 0.5 & 96.1 & 0.4 & 87.6 & 1.2 &  88.8 & 2.4 & 2.92 & 39.2 \\ 
\midrule 
\rowcolor{Gray!30}
BSL-Ens: S(B)\&S(J+B)\&V(J+B) & - & 3 & 90.7 & - & 95.7 & - & 87.3 & - & 86.9 & - & 4.30 & 57.8 \\ 
\ange{}-Ens: S(B)\&S(J+B)\&V(J+B) & - & 3 & 91.4 & 0.7 & \textbf{\textcolor{blue}{96.3}} & 0.6 & \textbf{\textcolor{red}{88.4}} & 1.1 & \textbf{\textcolor{blue}{89.1}} & 2.2 & 4.36 & 58.6 \\ 
\midrule 
\rowcolor{Gray!30}
BSL-Ens: S(J)\&S(B)\&S(J+B)\&V(J+B) & - & 4 & 90.9 & - & 95.9 & - & 87.5 & - & 87.2 & - & 5.72 & 76.8 \\ 
\ange{}-Ens: S(J)\&S(B)\&S(J+B)\&V(J+B) & - & 4 & \textbf{\textcolor{red}{91.6}} & 0.7 & \textbf{\textcolor{blue}{96.3}} & 0.4 & \textbf{\textcolor{blue}{88.2}} & 0.7 & \textbf{\textcolor{red}{89.2}} & 2.0 & 5.80 & 78.0 \\ 
\bottomrule
\end{tabular}
}
\label{tab:compare_with_sota}
\end{table*}
\begin{table}[t]
\centering
\caption{Evaluation results on ensembling with angular features. Ens is the ensembling. Jnt and Bon represent the joint and bone features respectively. The red bold number highlights the highest prediction accuracy. Acc$\uparrow$ is the improvement in accuracy. } 
% \vspace{-2mm}
\resizebox{0.48\textwidth}{!}{
\begin{tabular}{l|rr|rr}
\toprule 
%\rowcolor{Gray}
Features & Distance & Acc$\uparrow$ (\%) & Velocity & Acc$\uparrow$ (\%)\\ 
\midrule 
Ang & 81.97 & -- & 79.83 & -- \\ \midrule
\rowcolor{Gray!30}
Jnt & 81.90 & -- & 79.31 & -- \\ 
Ens: Jnt \& Ang & 83.53 & 1.63 & 83.81 & 4.5 \\ \midrule
\rowcolor{Gray!30}
Bon & 84.00 & -- & 80.32 & -- \\ 
Ens: Bon \& Ang & 86.47 & 2.47 & 86.13 & 5.81\\ 
\midrule 
\rowcolor{Gray!30}
Ens: Jnt+Bon & 86.22 & -- & 86.35 & --\\ 
Ens: Jnt+Bon \& Ang & \textcolor{red}{\textbf{87.13}} & 0.91 & 86.87 & 0.52\\ 
\bottomrule
% \hline
% \hline
\end{tabular}
}
\label{table:ang_feature_ens}
\end{table}

\begin{table*}[t]
\caption{A comparison of with/without angular features on the most confusing actions that may share similar motion trajectories.
The `Action' column shows the ground truth labels, and the `Similar Action' column shows the predictions from the model (with/without angular features).
The similar actions highlighted in orange demonstrate the change of predictions after employing angular features. 
The accuracy improvements highlighted in red are the substantially increased ones (Acc$\uparrow$ $\geq$ 10\%) due to using our angular features.
} 
\centering
% \vspace{-2mm}
\resizebox{\textwidth}{!}{
\begin{tabular}{l|rc|rrc}
\toprule 
%\rowcolor{Gray}
 & \multicolumn{2}{c|}{\textbf{Joint}} & \multicolumn{3}{c}{\textbf{Concatenation: Joint + Angular}} \\
\cline{2-3}\cline{4-6}
%\rowcolor{Gray}
\multirow{-2}{*}{\textbf{Action}} & Acc (\%) & Similar Action & Acc (\%) & Acc$\uparrow$ (\%) & Similar Action \\ 
\midrule 
\rowcolor{Gray!30} 
make victory sign & 18.48 & \textcolor{orange}{thumb up}   & 53.04 & \textcolor{red}{\textbf{34.57}} & \textcolor{orange}{make ok sign}   \\
staple book & 26.67 & \textcolor{orange}{staple book}   & 37.13 & \textcolor{red}{\textbf{10.46}} & \textcolor{orange}{cutting paper (using scissors)}   \\
\rowcolor{Gray!30}
writing & 28.41 & typing on a keyboard   & 48.90 & \textcolor{red}{\textbf{20.49}} & typing on a keyboard   \\
counting money & 48.47 & play magic cube   & 52.98 & 4.51 & play magic cube   \\
\rowcolor{Gray!30}
playing with phone/tablet & 48.82 & \textcolor{orange}{play magic cube}   & 59.64 & \textcolor{red}{\textbf{10.82}} & \textcolor{orange}{writing}   \\
wield knife towards other person & 49.52 & hit other person with something   & 62.50 & \textcolor{red}{\textbf{12.98}} & hit other person with something   \\
% \rowcolor{Gray!30}
% make ok sign & 54.48 & make victory sign   & 48.52 & -5.96 & make victory sign   \\
blow nose & 55.35 & yawn   & 59.65 & 4.30 & yawn   \\
\rowcolor{Gray!30}
fold paper & 56.57 & \textcolor{orange}{ball up paper}   & 62.78 & 6.22 & \textcolor{orange}{counting money}   \\
reading & 58.34 & \textcolor{orange}{cutting paper (using scissors)}   & 64.10 & 5.76 & \textcolor{orange}{writing}   \\
\rowcolor{Gray!30}
thumb up & 58.65 & make victory sign   & 72.35 & \textcolor{red}{\textbf{13.70}} & make victory sign   \\
yawn & 59.00 & hush (quite)   & 67.65 & 8.65 & hush (quite)   \\
\rowcolor{Gray!30}
snapping fingers & 59.10 & \textcolor{orange}{shake fist}   & 65.51 & 6.40 & \textcolor{orange}{make victory sign}   \\
open a box & 59.98 & \textcolor{orange}{fold paper}   & 71.60 & \textcolor{red}{\textbf{11.63}} & \textcolor{orange}{open bottle}   \\
% \rowcolor{Gray!30}
% play magic cube & 62.81 & \textcolor{orange}{counting money}   & 59.97 & -2.85 & playing with phone/tablet   \\
% cutting nails & 63.32 & playing with phone/tablet   & 63.09 & -0.23 & playing with phone/tablet   \\
\rowcolor{Gray!30}
pointing to something with finger & 64.58 & taking a selfie   & 79.71 & \textcolor{red}{\textbf{15.13}} & taking a selfie   \\
sneeze/cough & 64.58 & touch head (headache)   & 71.74 & 7.16 & touch head (headache)   \\
\rowcolor{Gray!30}
apply cream on hand back & 67.82 & \textcolor{orange}{open bottle}   & 72.30 & 4.48 & \textcolor{orange}{rub two hands together}   \\
cutting paper (using scissors) & 68.28 & staple book   & 70.16 & 1.87 & staple book   \\
\bottomrule 
\end{tabular}
}
\label{table:confusion}
\end{table*}

\begin{table*}[htbp]
\caption{
A comparison of the effect for improving action recognition by concatenating certain angular features to the joint representation. Each subtable is sorted by the increase in accuracy.  
The `Action' column shows the ground truth labels, and the `Similar Action' column shows the predictions from the model (with/without angular encoding).
} 
% \vspace{-2mm}
% Subtable 1
% \caption{Actions whose accuracy values are boosted by the local angular feature. }
\resizebox{\textwidth}{!}{
\begin{tabular}{c|l|rc|rrc}
\toprule 
%\rowcolor{Gray}
& & \multicolumn{2}{c|}{\textbf{Joint}} & \multicolumn{3}{c}{\textbf{Concatenation: Joint + Angular}} \\
\cline{3-4}\cline{5-7}
%\rowcolor{Gray}
& \multirow{-2}{*}{\textbf{Action}} & Acc (\%) & Similar Action & Acc (\%) & Acc$\uparrow$ (\%) & Similar Action \\ 
\midrule 
\multirow{7}{*}{\STAB{\rotatebox[origin=c]{90}{Static}}} 
% & make victory sign & 18.48 & thumb up   & 40.87 & 22.39 & make ok sign   \\
% & writing & 28.41 & typing on a keyboard   & 45.96 & 17.54 & typing on a keyboard   \\
& wear a shoe & 70.43 & take off a shoe   & 86.08 & 15.65 & take off a shoe   \\
% & playing with phone/tablet & 48.82 & play magic cube   & 64.36 & 15.55 & writing   \\
& punching/slapping other person & 72.36 & hit other person with something   & 85.40 & 13.04 & hit other person with something   \\
& thumb up & 58.65 & make victory sign   & 71.13 & 12.48 & make victory sign   \\
& pointing to something with finger & 64.58 & taking a selfie   & 75.72 & 11.14 & taking a selfie   \\
& wield knife towards other person & 49.52 & hit other person with something   & 60.24 & 10.72 & hit other person with something   \\
& fold paper & 56.57 & ball up paper   & 66.61 & 10.04 & counting money   \\
& open a box & 59.98 & fold paper   & 68.47 & 8.49 & fold paper   \\
\midrule
\multirow{7}{*}{\STAB{\rotatebox[origin=c]{90}{Velocity}}} 
& cutting paper (using scissors) & 27.27 & staple book   & 45.90 & 18.63 & staple book   \\
& playing with phone/tablet & 39.73 & writing   & 57.45 & 17.73 & typing on a keyboard   \\
% & make ok sign & 27.17 & make ok sign   & 40.87 & 13.70 & make victory sign   \\
& drink water & 72.72 & brushing teeth   & 83.94 & 11.22 & brushing teeth   \\
& play magic cube & 45.50 & counting money   & 56.64 & 11.14 & counting money   \\
& reading & 48.82 & writing   & 59.71 & 10.89 & writing   \\
& typing on a keyboard & 56.45 & writing   & 67.27 & 10.82 & writing   \\
% & touch head (headache) & 65.67 & brushing teeth   & 75.72 & 10.06 & brushing teeth   \\
% & take off a shoe & 76.74 & wear a shoe   & 86.13 & 9.39 & wear a shoe   \\
& wipe face & 75.09 & touch head (headache)   & 83.70 & 8.61 & touch head (headache)   \\
\bottomrule % Table 2
\toprule 
& & \multicolumn{2}{c|}{\textbf{Joint}} & \multicolumn{3}{c}{\textbf{Concatenation: Joint + Angular}} \\
\cline{3-4}\cline{5-7}
%\rowcolor{Gray}
& \multirow{-2}{*}{\textbf{Action}} & Acc (\%) & Similar Action & Acc (\%) & Acc$\uparrow$ (\%) & Similar Action \\ 
\midrule 
\multirow{9}{*}{\STAB{\rotatebox[origin=c]{90}{Static}}} 
& make victory sign & 18.48 & thumb up   & 40.35 & 21.87 & make ok sign   \\
% & writing & 28.41 & typing on a keyboard   & 49.63 & 21.22 & typing on a keyboard   \\
& playing with phone/tablet & 48.82 & play magic cube   & 68.36 & 19.55 & staple book   \\
& wield knife towards other person & 49.52 & hit other person with something   & 65.80 & 16.28 & hit other person with something   \\
& wear a shoe & 70.43 & take off a shoe   & 85.35 & 14.92 & take off a shoe   \\
& take off a shoe & 70.90 & wear a shoe   & 85.40 & 14.50 & wear a shoe   \\
& punching/slapping other person & 72.36 & hit other person with something   & 83.21 & 10.85 & hit other person with something   \\
& yawn & 59.00 & hush (quite)   & 69.57 & 10.57 & blow nose   \\
& pointing to something with finger & 64.58 & taking a selfie   & 75.00 & 10.42 & taking a selfie   \\
& fold paper & 56.57 & ball up paper   & 66.09 & 9.52 & ball up paper   \\
\midrule 
\multirow{7}{*}{\STAB{\rotatebox[origin=c]{90}{Velocity}}} 
& cutting paper (using scissors) & 27.27 & staple book   & 58.12 & 30.84 & staple book   \\
& playing with phone/tablet & 39.73 & writing   & 56.73 & 17.00 & staple book   \\
& make ok sign & 27.17 & make ok sign   & 43.65 & 16.48 & make victory sign   \\
& play magic cube & 45.50 & counting money   & 61.19 & 15.69 & counting money   \\
& drink water & 72.72 & brushing teeth   & 87.96 & 15.23 & brushing teeth   \\
& typing on a keyboard & 56.45 & writing   & 70.18 & 13.73 & writing   \\
& touch head (headache) & 65.67 & brushing teeth   & 77.90 & 12.23 & drink water   \\
% & pointing to something with finger & 60.96 & taking a selfie   & 72.83 & 11.87 & taking a selfie   \\
% & reading & 48.82 & writing   & 57.88 & 9.06 & writing   \\
% & wipe face & 75.09 & touch head (headache)   & 84.06 & 8.97 & touch head (headache)   \\
\bottomrule
\toprule % Table 3 Part
%\rowcolor{Gray}
& & \multicolumn{2}{c|}{\textbf{Joint}} & \multicolumn{3}{c}{\textbf{Concatenation: Joint + Angular}} \\
\cline{3-4}\cline{5-7}
%\rowcolor{Gray}
& \multirow{-2}{*}{\textbf{Action}} & Acc (\%) & Similar Action & Acc (\%) & Acc$\uparrow$ (\%) & Similar Action \\ 
\midrule
\multirow{7}{*}{\STAB{\rotatebox[origin=c]{90}{Static}}} 
% & writing & 28.41 & typing on a keyboard   & 49.63 & 21.22 & typing on a keyboard   \\
& make victory sign & 18.48 & thumb up   & 37.39 & 18.91 & make ok sign   \\
& open a box & 59.98 & fold paper   & 74.56 & 14.59 & open bottle   \\
& wear a shoe & 70.43 & take off a shoe   & 84.98 & 14.55 & take off a shoe   \\
& wield knife towards other person & 49.52 & hit other person with something   & 63.37 & 13.85 & hit other person with something   \\
& pointing to something with finger & 64.58 & taking a selfie   & 77.17 & 12.59 & taking a selfie   \\
% & nod head/bow & 85.23 & nausea or vomiting condition   & 96.01 & 10.78 & take off a shoe   \\
& take off a shoe & 70.90 & wear a shoe   & 79.93 & 9.03 & wear a shoe   \\
% & apply cream on hand back & 67.82 & open bottle   & 76.48 & 8.67 & rub two hands together   \\
& thumb down & 75.52 & thumb up   & 83.48 & 7.96 & thumb up   \\
\midrule
\multirow{7}{*}{\STAB{\rotatebox[origin=c]{90}{Velocity}}}
& cutting paper (using scissors) & 27.27 & staple book   & 59.34 & 32.06 & staple book   \\
& playing with phone/tablet & 39.73 & writing   & 71.27 & 31.55 & typing on a keyboard   \\
& play magic cube & 45.50 & counting money   & 64.86 & 19.36 & counting money   \\
& typing on a keyboard & 56.45 & writing   & 72.00 & 15.55 & writing   \\
% & thumb up & 53.43 & make victory sign   & 66.61 & 13.17 & make victory sign   \\
% & make ok sign & 27.17 & make ok sign   & 40.17 & 13.00 & make victory sign   \\
& pointing to something with finger & 60.96 & taking a selfie   & 73.55 & 12.59 & taking a selfie   \\
& drink water & 72.72 & brushing teeth   & 85.04 & 12.31 & brushing teeth   \\
& open a box & 56.84 & open bottle   & 68.82 & 11.98 & open bottle   \\
% & reading & 48.82 & writing   & 59.71 & 10.89 & writing   \\
\bottomrule 
\toprule % Table 4 Finger
%\rowcolor{Gray}
& & \multicolumn{2}{c|}{\textbf{Joint}} & \multicolumn{3}{c}{\textbf{Concatenation: Joint + Angular}} \\
\cline{3-4}\cline{5-7}
%\rowcolor{Gray}
& \multirow{-2}{*}{\textbf{Action}} & Acc (\%) & Similar Action & Acc (\%) & Acc$\uparrow$ (\%) & Similar Action \\ 
\midrule
\multirow{7}{*}{\STAB{\rotatebox[origin=c]{90}{Static}}} 
% & writing & 28.41 & typing on a keyboard   & 53.31 & 24.90 & typing on a keyboard   \\
& make victory sign & 18.48 & thumb up   & 39.48 & 21.00 & make ok sign   \\
% & wear a shoe & 70.43 & take off a shoe   & 89.74 & 19.32 & take off a shoe   \\
& wield knife towards other person & 49.52 & hit other person with something   & 63.72 & 14.19 & hit other person with something   \\
& playing with phone/tablet & 48.82 & play magic cube   & 61.45 & 12.64 & play magic cube   \\
& punching/slapping other person & 72.36 & hit other person with something   & 82.85 & 10.49 & wield knife towards other person   \\
& fold paper & 56.57 & ball up paper   & 65.57 & 9.00 & ball up paper   \\
% & touch head (headache) & 75.45 & drink water   & 84.06 & 8.61 & blow nose   \\
& play magic cube & 62.81 & counting money   & 71.15 & 8.34 & playing with phone/tablet   \\
& side kick & 84.89 & kicking something   & 93.21 & 8.32 & kicking something   \\
\midrule
\multirow{7}{*}{\STAB{\rotatebox[origin=c]{90}{Velocity}}}
& playing with phone/tablet & 39.73 & writing   & 66.18 & 26.45 & typing on a keyboard   \\
& cutting paper (using scissors) & 27.27 & staple book   & 53.40 & 26.13 & staple book   \\
& play magic cube & 45.50 & counting money   & 64.86 & 19.36 & counting money   \\
& typing on a keyboard & 56.45 & writing   & 74.18 & 17.73 & writing   \\
& pointing to something with finger & 60.96 & taking a selfie   & 78.26 & 17.30 & taking a selfie   \\
& drink water & 72.72 & brushing teeth   & 85.04 & 12.31 & brushing teeth   \\
% & make ok sign & 27.17 & make ok sign   & 36.70 & 9.52 & make victory sign   \\
% & throw & 82.27 & tennis bat swing   & 91.64 & 9.36 & tennis bat swing   \\
% & blow nose & 52.04 & yawn   & 61.39 & 9.35 & hush (quite)   \\
& nausea or vomiting condition & 75.36 & touch chest (stomachache/heart pain)   & 84.36 & 9.00 & touch chest (stomachache/heart pain)   \\
\bottomrule 
\end{tabular}
}
\label{table:diff_angle_analysis}
\end{table*}

\begin{table}[th]
    \caption{Comparison of recognition performance between MSGCN and MSG3D. MSG3D has higher accuracy, more parameters, and a longer running time. GFlops stands for the floating-point operations performed by a model, which is the number of multiply-add operations that a model performs. }
    \centering
    \resizebox{0.48\textwidth}{!}{
    \begin{tabular}{l|c|c|c|c}
    \toprule
        Architecture & \makecell{Static: \\Jnt+Bon+Ang} & \makecell{Velocity: \\Jnt+Bon+Ang} & \# Params & GFlops \\
        \midrule
        \rowcolor{Gray!30}
        MSGCN+Ang & 84.6 & 83.2 & 1.46 & 19.6 \\
        MSG3D+Ang & 86.2 & 83.6 & 3.24 & 50.0 \\
    \bottomrule
    \end{tabular}
    }
    \label{tab:msg3d_compare}
\end{table}

\section{Related Work}
Many of the earliest attempts at skeleton-based action recognition encoded all human body joint coordinates in each frame into a feature vector for pattern learning~\cite{lei2017,lei_tip2019}. These models rarely explored the internal dependencies between body joints, resulting in missing rich information about actions. Kernel-based methods have also been proposed for action recognition~\cite{koniusz2016tensor, koniusz2020tensor}. 

Later, as deep learning became a standard choice in video processing~\cite{liu2021invertible,anwar2020densely} and understanding~\cite{li2020transferring,li2020word}, RGB-based videos started to tackle action recognition. However, they suffer from problems in domain adaptation~\cite{zhang2020clarinet, fang2020open, zhong2020does} since they have varying backgrounds with different textures of subjects. On the other hand, skeleton data has relatively fewer issues with domain adaptation. Convolutional neural networks (CNNs) were introduced to tackle skeleton-based action recognition and achieved an improvement~\cite{lei_iccv19}. However, CNNs are designed for grid-based data and are not suitable for graph data since they cannot leverage the topology of a graph. 

Recently, deep graph neural networks are accumulating attention~\cite{micheli2009neural, yao2022multi, li2020fast, wang2020haar}. Graph neural networks also started to attract attention in skeleton recognition. In GCN-based models, a skeleton is treated as a graph, with joints as nodes and bones as edges. An early application was ST-GCN~\cite{yan2018spatial}, using graph convolution to aggregate joint features spatially and convolving consecutive frames along the temporal axis. Subsequently, AS-GCN~\cite{Li_2019_CVPR} was proposed to further improve the spatial feature aggregation via the learnable adjacency matrix instead of using the skeleton as a fixed graph.  AGC-LSTM~\cite{Si_2019_CVPR} learned long-range temporal dependencies, using LSTM as a backbone, and changed every gate operation from the original fully connected layer to a graph convolution layer, making better use of the skeleton topological information.
2s-AGCN~\cite{shi2019two} made two major contributions: (a) applying a learnable residual mask to the adjacency matrix of the graph convolution, making the skeleton's topology more flexible; (b) proposing a second-order feature, the difference between the coordinates of two adjacent joints, to act as the bone information. An ensemble of two models, trained with the joint and bone features, substantially improved the classification accuracy. More graph convolution techniques have been proposed in skeleton-based action recognition, such as SGN~\cite{zhang2020semantics} and Shift-GCN~\cite{cheng2020skeleton}, employing self-attention and shift convolution respectively. Recently, MS-G3D~\cite{liu2020disentangling} achieved high results by proposing graph 3D convolutions to aggregate features within a window of consecutive frames. However, 3D convolutions demand a long running time.

In more recent times, Qin \etal proposed some self-attention models that dynamically optimize the graph structure~\cite{qin2022efficient}. Xu \etal designed a pure CNN architecture that more effectively captures the topological information~\cite{xu2022topology}. Memmesheimer \etal study the one-shot problem of skeleton-based action recognition~\cite{memmesheimer2022skeleton}. They apply the metric learning setting and map the problem to a nearest-neighbor search in a set of activity reference samples. Wang \etal studied the adversarial attack problem in skeleton-based action recognition~\cite{wang2021understanding}. They investigated a perceptual loss that ensures the imperceptibility of the attack. Diao \etal investigated the black-box attack on skeleton-based action recognition~\cite{diao2021basar}. They proposed an attack mechanism called BASKR and showed that the adversarial attack is a threat and on-manifold adversarial samples are common for skeletal motions. 

All the existing methods suffer from low accuracy in discriminating actions sharing similar motion trajectories. This motivates us to seek a new encoding to facilitate the model differentiating two confusing actions. Some works show angle features similar to the local feature presented in this paper~\cite{hu2022skeleton,yadav2022skeleton}. On the other hand, we propose a collection of angular encoding forms. Each category consists of further subcategories. Different categories of angular encoding are designed to capture motion features of distinct kinematic body parts.

\section{Angular Feature Representation}
\subsection{Angular Encoding}

We propose using third-order features, which measure the angle between three body joints to depict the relative movements between body parts in skeleton-based action recognition. Given three joints $u$, $w_1$ and $w_2$, where $u$ is the target joint to calculate the angular features and $w_1$ and $w_2$ are endpoints in the skeleton, 
$\vec{b}_{uw_i}$ denotes the vector from joint $u$ to $w_i$ ($i = 1, 2$), we have $\vec{b}_{uw_i} = (x_{w_i} - x_u, y_{w_i} - y_u, z_{w_i} - z_u)$, where $(x_k,y_k,z_k)$ represent the coordinates of joint $k$ ($k=u, w_1, w_2$). We define two kinds of angular features. 

\textit{Static Angular Encoding}:
suppose $\theta$ is the angle between $\vec{b}_{uw_1}$ and $\vec{b}_{uw_2}$; we define the \emph{static angular encoding} $d_a(u)$ for joint $u$ as 
\begin{align}
    d_a(u)  
    = 
    \begin{cases}
      1 - \cos \theta = 
      1 - \frac{ \vec{b}_{uw_1}\cdot\vec{b}_{uw_2} }{ |\vec{b}_{uw_1}| |\vec{b}_{uw_2}| }  & \hspace{-2mm} \text{if $u \ne w_1$, $u \ne w_2$,}\\
      0 & \hspace{-5mm} \text{if $u = w_1$ or $u = w_2$.}
    \end{cases}
\end{align}

Note that $w_1$ and $w_2$ do not need to be adjacent nodes of $u$. The feature value increases monotonically as $\theta$ goes from $0$ to $\pi$ radians. 
In contrast to the first-order features, representing the coordinate of a joint, and the second-order features, representing the lengths and directions of bones, these third-order features focus more on motions and are invariant to the scale of human subjects. 

\textit{Velocity Angular Encoding}: the temporal differences of the angular features between consecutive frames, \ie
\begin{equation}
    v_a^{(t+1)}(u) = d_a^{(t+1)}(u) - d_a^{t}(u),
\end{equation}
where $v_a^{(t+1)}(u)$ is the angular velocity of joint $u$ at frame $(t+1)$, describing the dynamic changes of angles. The angular encoding is a third-order feature. Taking the velocity of these third-order features further increases the order. Hence, these velocity angular features enable an action recognizer to capture fourth-order information of motion sequences. 

However, we face a computational challenge when we attempt to exploit these angular features: if we use all possible angles, \ie{} all possible combinations of $u$, $w_1$ and $w_2$, the computational complexity is $O(N^3 T)$, where $N$ and $T$ respectively represent the number of joints and frames.
Instead, we manually define sets of angles that seem likely to facilitate distinguishing actions without drastically increasing computational cost. In the rest of this section, we present the four categories of angles considered in this work.

\textbf{(a) Locally-Defined Angles.} As illustrated in Figure~\ref{fig:skeletons}(a), a locally-defined angle is measured between a joint and its two adjacent neighbors. If the target joint has only one adjacent joint, we set its angular feature to zero. When a joint has more than two adjacent joints, we choose the most active two. For example, we use the two shoulders instead of the head and belly for the neck joint since the latter rarely move. 
These angles can capture relative motions between two bones. 

\textbf{(b) Center-Oriented Angles.} A center-oriented angle measures the angular distance between a target joint and two body center joints representing the neck and pelvis. As in \autoref{fig:skeletons}(b), given a target joint, we use two center-oriented angles: 1) neck-target-pelvis, dubbed as unfixed-axis, and 2) neck-pelvis-target, dubbed as fixed-axis. For the joints representing the neck and pelvis, we set their angular features to zero.
Center-oriented angles measure the relative position between a target joint and the body center joints. For example, given an elbow as a target joint moving away horizontally from the body center, the unfixed-axis angle decreases while the fixed-axis angle increases. 

\textbf{(c) Pair-Based Angles.} Pair-based angles measure the angle between a target joint and four pairs of endpoints: hands, elbows, knees, and feet, as illustrated in Figure~\ref{fig:skeletons}(c).
If the target joint is one of the endpoints, we set the feature value to zero. We select these four pairs due to their importance in performing actions. The pair-based angles are beneficial for recognizing object-related actions. For example, when a person is holding a box, the angle between a target joint and hands can indicate the box's size.

\textbf{(d) Finger-Based Angles.} 
Fingers are actively involved in human actions.
When the skeleton of each hand has finger joints,
we include more detailed finger-based angles to incorporate them.
As demonstrated in Figure~\ref{fig:skeletons}(d), the two joints corresponding to fingers are selected as the anchor endpoints of an angle. The finger-based angles can indirectly depict gestures. For instance, 
an angle with a wrist as the root and a hand tip as well as a thumb as two endpoints can reflect the degree of hand opening. 

\subsection{Our Backbone Architecture}
\label{sec: angnet}
The overall network architecture is illustrated in \autoref{fig:net_architecture}. 
Three different features are extracted from the skeleton and input into the stack of three spatial-temporal blocks (STBs). Then, the output passes sequentially to a global average pooling, a fully connected layer, and then a softmax layer for action classification. We use a simplified version of MS-G3D \cite{liu2020disentangling} as the backbone of our model. 
For simplification, we remove their heavy graph 3D convolution (G3D) modules, weighing the performance gain against the computational cost. 
We call the resulting system \ourmodel{}. Note that our proposed angular features are independent of the choice of the backbone. 

We extract the joint, bone, and angular features from every action video. For the bone feature, if a joint has more than one adjacent node, we choose the joint closer to the body's center. So, given an elbow joint, we use the vector from the elbow to the shoulder rather than the vector from the elbow to the wrist. For the angle, we extract seven or nine angular features (without/with finger-based angles) for every joint, constituting seven or nine channels of features. Eventually, for each action, we construct a feature tensor $X \in \mathbb{R}^{C \times T \times V \times M}$, where $C$, $T$, $V$ and $M$ respectively correspond to the numbers of channels, frames, joints, and participants (the persons conducting actions). We test various combinations of the joint, bone, and angular features in the experiments. 

Each STB, as exhibited in \autoref{fig:net_architecture}(b), comprises a spatial multiscale graph convolution (SMGC) unit and three temporal multiscale convolution (TMC) units. The details of these components are illustrated as follows. 

The SMGC unit, as shown in \autoref{fig:net_architecture}(c), consists of a parallel combination of graph convolutional layers. The adjacency matrix of graph convolutions results from the summation of a powered adjacency matrix $A^k$ and a learnable mask $A^k_{mask}$. 
\textit{Powered adjacency matrices:}
To prevent over-smoothing, we avoid sequentially stacking multiple graph convolutional layers to make the network deep. Following~\cite{liu2020disentangling}, to create graph convolutional layers with different sizes of receptive fields, we directly use the powers of the adjacency matrix $A^{k}$ instead of $A$ itself to aggregate the multi-hop neighbor information. Thus, $A^{k}_{i,j} = 1$ indicates the existence of a path between joint $i$ and $j$ within $k$-hops. 
We feed the input into $K$ graph convolution branches with different receptive fields. $K$ is no more than the longest path within the skeleton graph. 
\textit{Learnable masks:} Using the skeleton as a fixed graph cannot capture the non-physical dependencies among joints. For example, two hands may always perform actions in conjunction, whereas they are not physically connected in a skeleton.
To infer the latent dependencies among joints, following~\cite{shi2019two}, we apply learnable masks to the adjacency matrices.

The TMC unit, shown in \autoref{fig:net_architecture}(d), consists of seven parallel temporal convolutional branches. 
Each branch starts with a $1 \times 1$ convolution to aggregate features between different channels. 
The functions of different branches diverge as the input passes forward, which can be divided into four groups. 
In detail: (a) \textit{Extracting multiscale temporal features}: the group contains four $3 \times 1$ temporal convolutions, applying four different dilations to obtain multiscale temporal receptive fields. (b) \textit{Processing features within the current frame}: This group only has one $1 \times 1$ to concentrate features within a single frame. (c) \textit{Emphasizing the most salient information within the consecutive frames}: The group ends with a $3 \times 1$ max-pooling layer to draw the most important features. (d) \textit{Preserving Gradient}: The final group incorporates a residual path to preserve gradients during back-propagation~\cite{chen2017dual}. 

% Visualization of actions 
\begin{figure*}[t]
\centering
% fold paper
\begin{subfigure}{.99\textwidth}
  \centering
  \includegraphics[width=.13\linewidth]{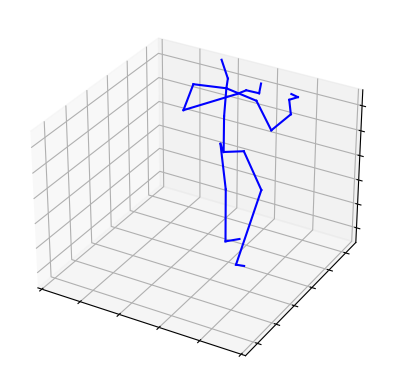}
  \includegraphics[width=.13\linewidth]{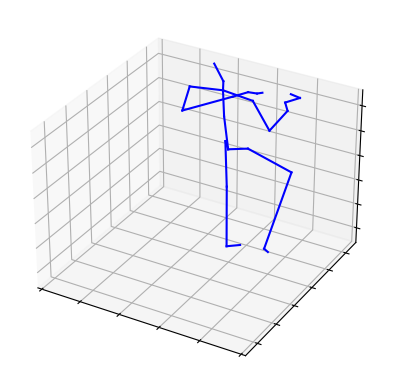}
  \includegraphics[width=.13\linewidth]{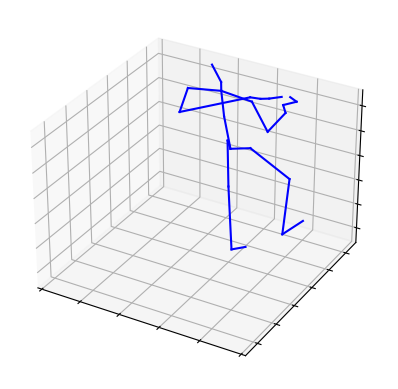}
  \includegraphics[width=.13\linewidth]{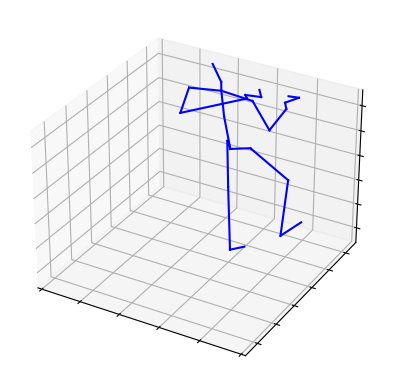}
  \includegraphics[width=.13\linewidth]{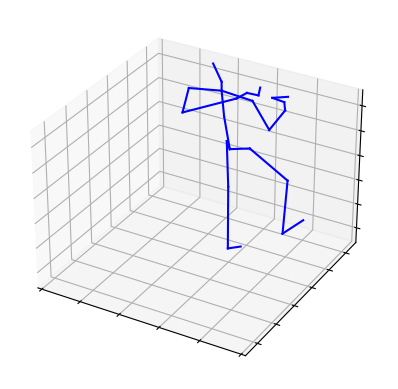}
  \includegraphics[width=.13\linewidth]{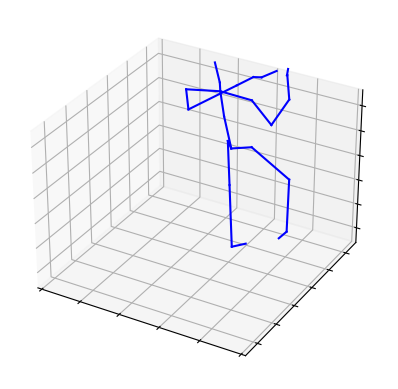}
  \includegraphics[width=.13\linewidth]{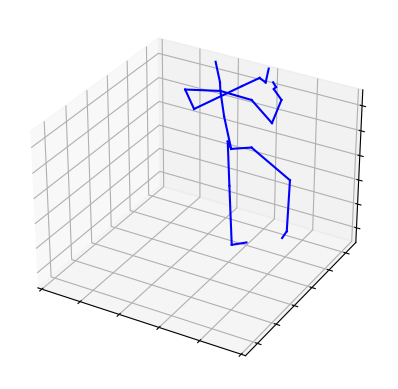}
  \caption{Folding paper. }
\end{subfigure}
% blow nose 
\begin{subfigure}{.99\textwidth}
  \centering
  \includegraphics[width=.13\linewidth]{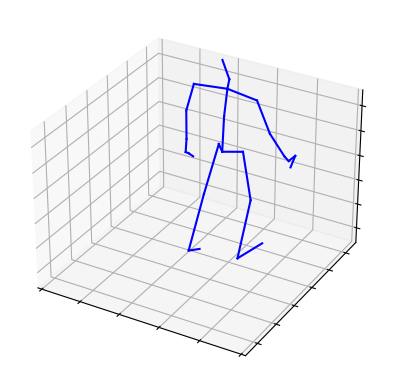}
  \includegraphics[width=.13\linewidth]{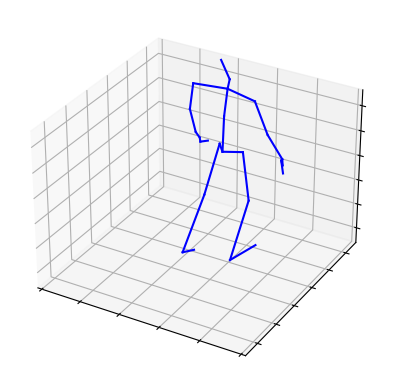}
  \includegraphics[width=.13\linewidth]{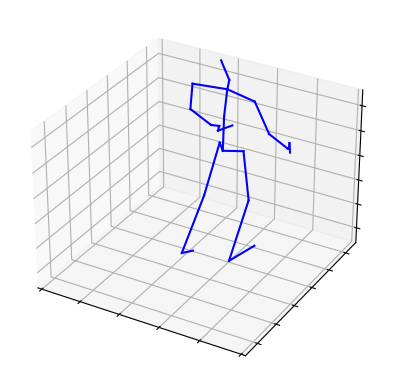}
  \includegraphics[width=.13\linewidth]{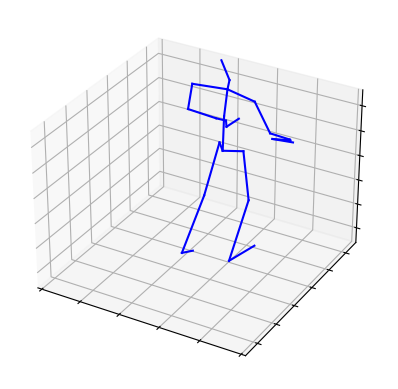}
  \includegraphics[width=.13\linewidth]{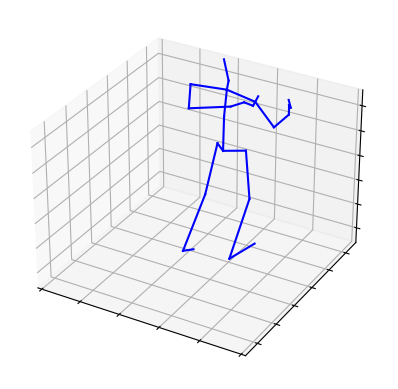}
  \includegraphics[width=.13\linewidth]{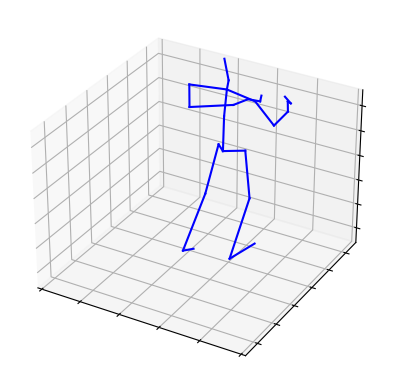}
  \includegraphics[width=.13\linewidth]{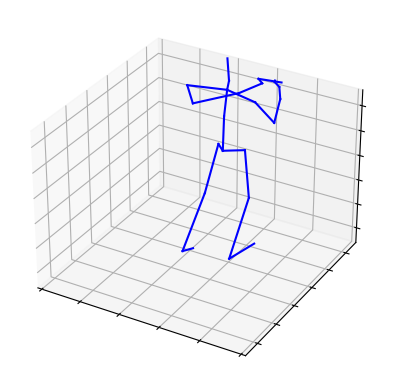}
  \caption{Counting money. }
\end{subfigure}
% snapping finger
\begin{subfigure}{.99\textwidth}
  \centering
  \includegraphics[width=.13\linewidth]{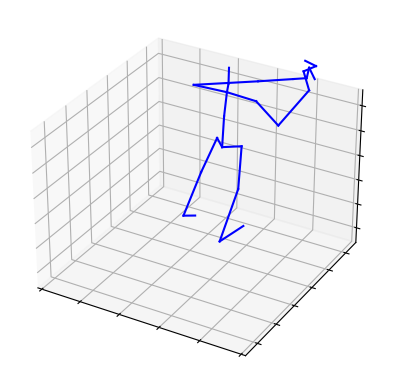}
  \includegraphics[width=.13\linewidth]{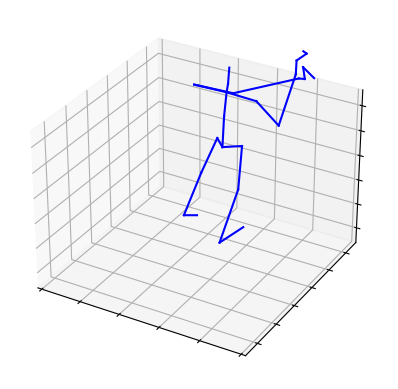}
  \includegraphics[width=.13\linewidth]{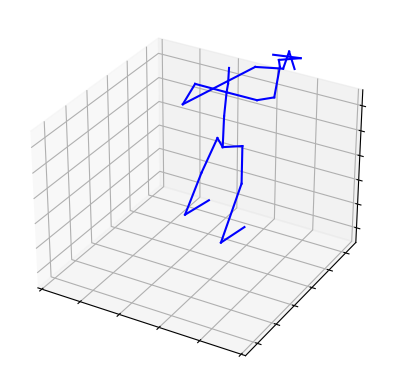}
  \includegraphics[width=.13\linewidth]{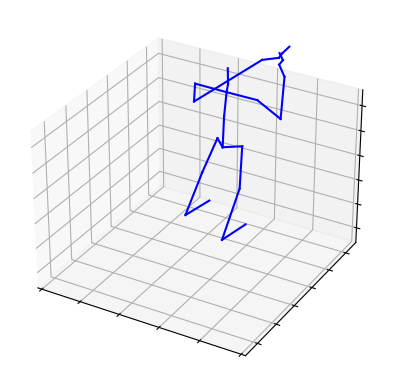}
  \includegraphics[width=.13\linewidth]{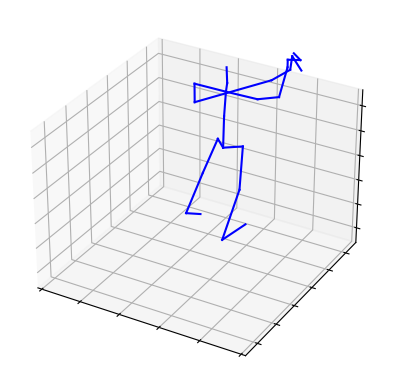}
  \includegraphics[width=.13\linewidth]{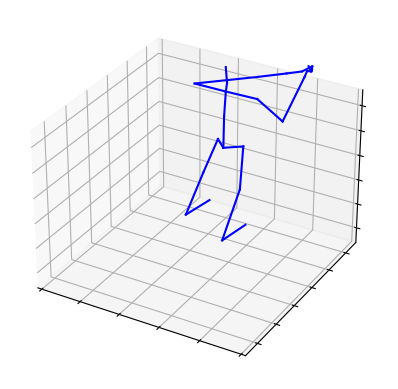}
  \includegraphics[width=.13\linewidth]{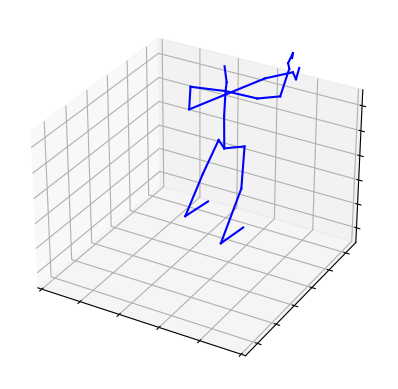}
  \caption{Reading. }
\end{subfigure}
% yawn 
\begin{subfigure}{.99\textwidth}
  \centering
  \includegraphics[width=.13\linewidth]{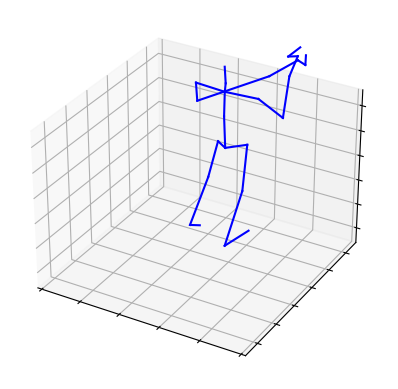}
  \includegraphics[width=.13\linewidth]{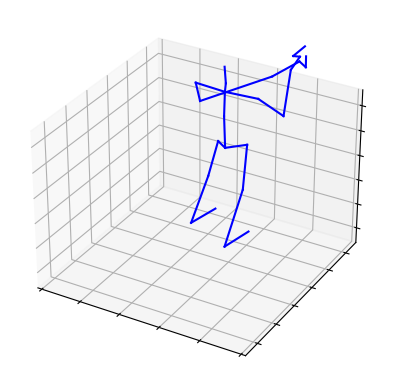}
  \includegraphics[width=.13\linewidth]{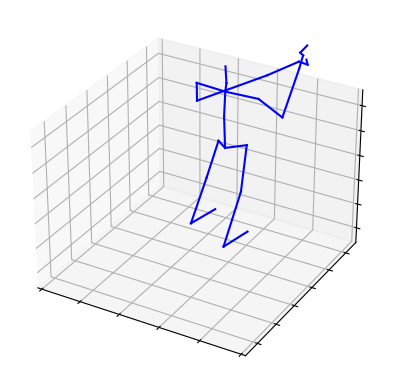}
  \includegraphics[width=.13\linewidth]{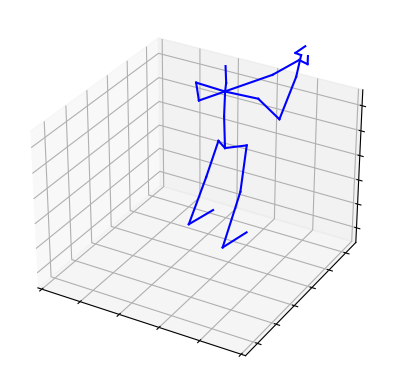}
  \includegraphics[width=.13\linewidth]{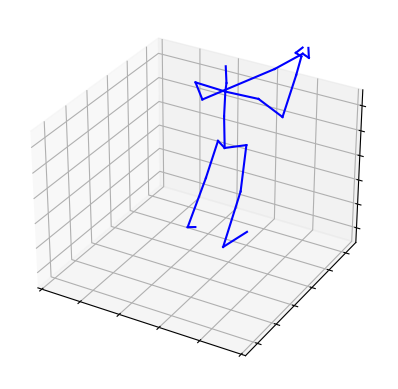}
  \includegraphics[width=.13\linewidth]{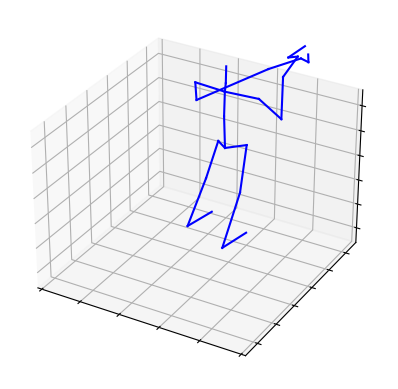}
  \includegraphics[width=.13\linewidth]{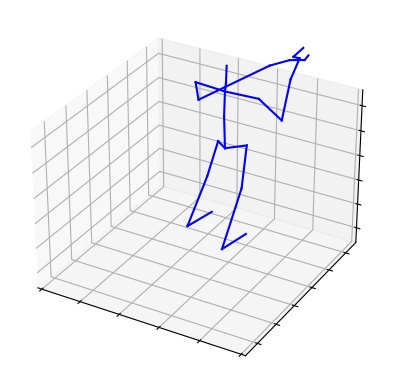}
  \caption{Writing. }
\end{subfigure}
\caption{
Visualization examples of confusing actions. The action that the network gets most confused about has changed after employing angular encoding as a part of input features. 
}
\label{fig:visual_action_sequence}
\vspace{-2mm}
\end{figure*}
\begin{figure}[htbp]
    \centering
    \includegraphics[width=0.95\linewidth]{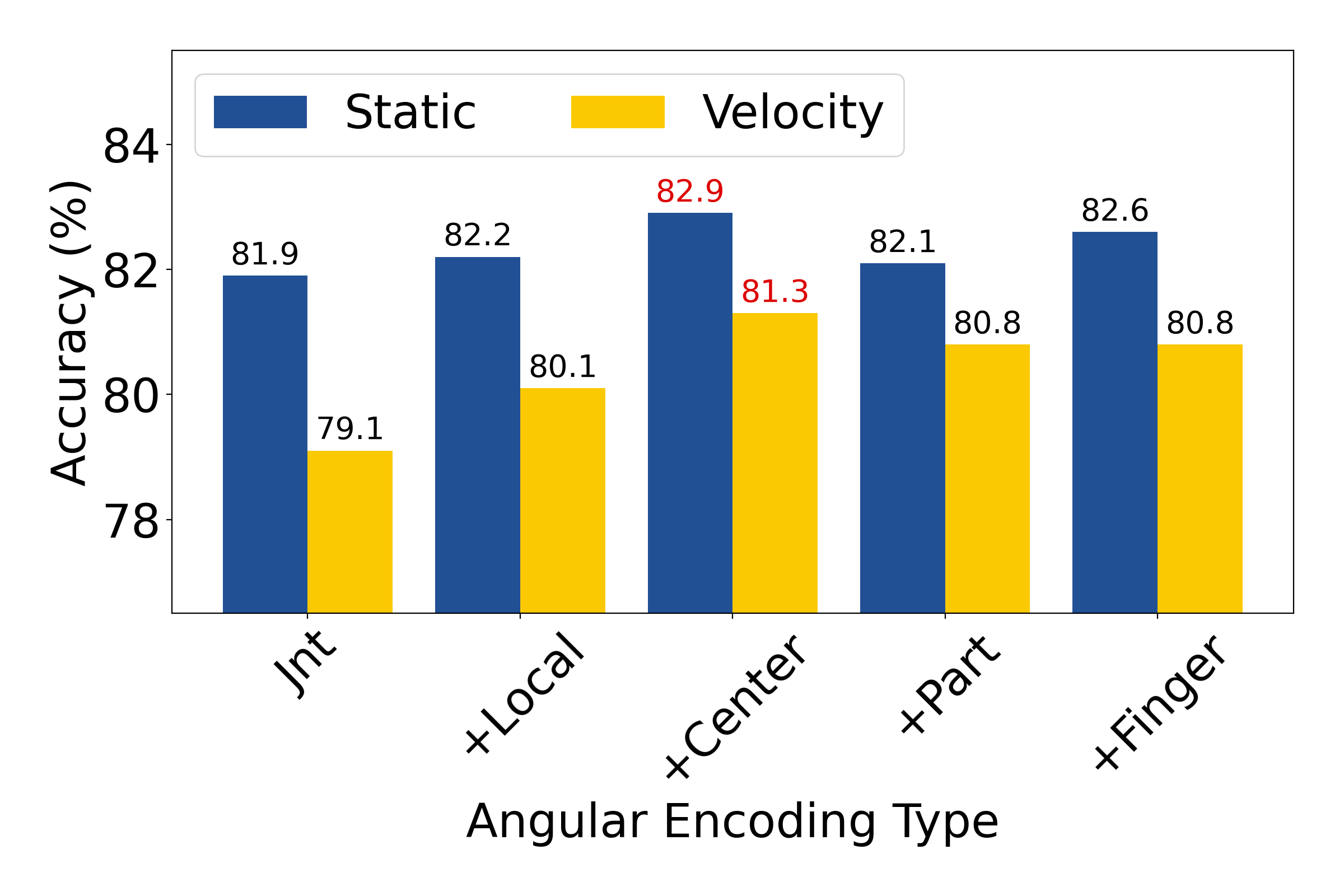}
    \caption{Accuracy of recognizing skeleton-based actions using the multi-scale GCN with different types of angular encoding. Both static and velocity domains are considered. The best accuracy of each domain is highlighted in red. }
    \label{fig:citation_network_bar_chart}
\end{figure}
\begin{figure*}[th]
\centering
\includegraphics[width=.45\linewidth]{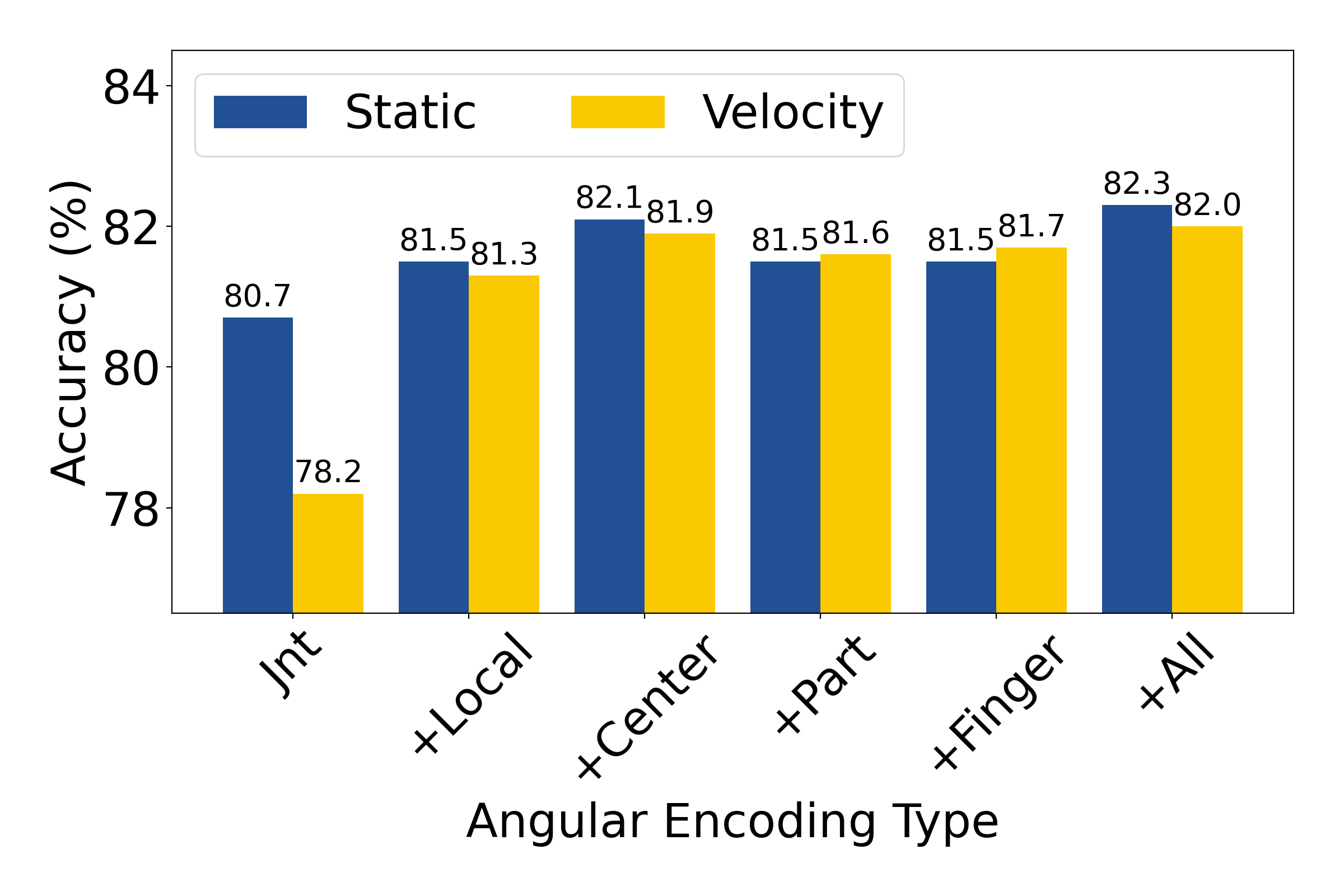}
\includegraphics[width=.45\linewidth]{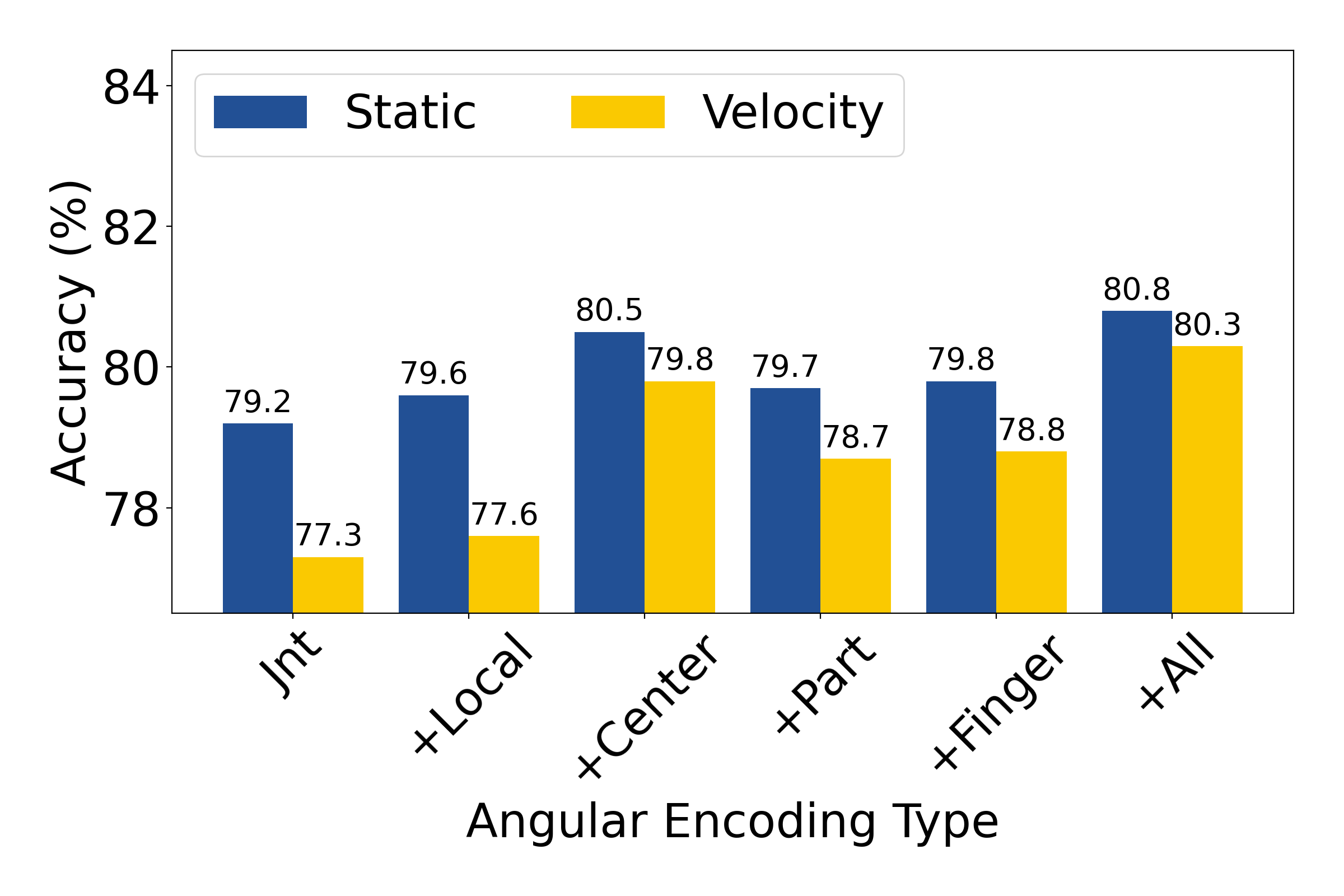} 
\caption{Accuracy of recognizing skeleton-based actions using DecoupleGCN (left) and ShiftGCN (right) with different types of angular encoding. Both static and velocity domains are considered. 
The column All represents concatenating all types of angular encoding. }
% \vspace{-4mm}
\label{fig:diff_backbone}
\end{figure*}

\section{Experiments}

\subsection{Datasets} 
\textbf{NTU60}~\cite{shahroudy2016ntu}.
NTU60 is a widely-used benchmark dataset for skeleton-based action recognition, incorporating 56,000 videos. The action videos were collected in a laboratory environment, resulting in accurately extracted skeletons. Nonetheless, recognizing actions from these skeletons is still challenging due to five aspects: (1) the skeletons are captured from different viewpoints; (2) the skeleton sizes of subjects vary; (3) so do their speeds of action; (4) different actions can have similar motion trajectories; (5) there are limited joints to portray hand actions in detail. 

\noindent
\textbf{NTU120}~\cite{liu2019ntu}.
% .NTU120 
NTU120 is an extension of NTU60. It uses more camera positions and angles, as well as a larger number of performing subjects, leading to 113,945 videos. 

\subsection{Experimental Setups}
We train deep learning models on four NVIDIA 2080-Ti GPUs and use PyTorch as our deep learning framework to compute the angular encoding. Furthermore, we apply stochastic gradient descent (SGD) with momentum 0.9 as the optimizer. The training epochs for NTU60 and NTU120 are set to 55 and 60, respectively, with learning rates decaying to 0.1 of the original value at epochs 35, 45, and 55. We follow \cite{shi2019skeleton} in normalizing, translating each skeleton, and padding all clips to 300 frames via repeating the action sequences. The training loss function is cross-entropy~\cite{qin2019rethinking}.

\begin{table}[t]
\centering
\caption{Independently evaluation of angular encoding for each category. XSub and XView represent cross-subject and cross-view. XSet means cross-setup. } 
% \vspace{-2mm}
\resizebox{0.48\textwidth}{!}{
\begin{tabular}{l|rr|rr}
\toprule 
%\rowcolor{Gray}
Angular Types & \makecell[r]{NTU60\\XSub} & \makecell[r]{NTU60\\XView} & \makecell[r]{NTU120\\XSub} & \makecell[r]{NTU120\\XSet}\\ 
\midrule 
No angular encoding & 87.2 & 93.7 & 81.9 & 83.5 \\ \midrule
\rowcolor{Gray!30}
With local & 87.9 & 94.1 & 82.8 & 83.5 \\ 
With center-based & 88.4 & 94.3 & 83.0 & 83.7 \\ \midrule
\rowcolor{Gray!30}
With pair-based & 87.8 & 94.2 & 82.4 & 83.5 \\ 
With finger-based & 88.0 & 94.1 & 82.7 & 83.6\\ \midrule
Concatenating all & 88.7 & 94.5 & 83.2 & 83.7\\ 
\bottomrule
% \hline
% \hline
\end{tabular}
}
\label{table:eval_each_ang}
\end{table}

\subsection{Ablation Studies}
There are two possible approaches for using angular features: (a) simply concatenate our proposed angular features with the existing joint, bone, or both features, and then train the model; (b) feed the angular features into our model and ensemble it with other models that are trained using joint, bone or both features to predict the action label. We study the differences between these approaches. We report the results in \autoref{tab:compare_with_sota}, including using different settings of both NTU and NTU120. To reduce clutter, we use the results of the cross-subject setting of NTU120 for ablation studies. 
We denote the accuracy without angular encoding with baseline (BSL). AGE means to concatenate the original feature with angular encoding. The suffix -S (in BSL-S and AGE-S) and -V (in BSL-V and AGE-V) represent feeding the static and velocity feature, respectively.

\textbf{Concatenating with Angular Features}. 
Here, we study the effects of concatenating angular features with others. We first obtain the accuracy of three models trained with three feature types, \ie{} the joint, bone, and a concatenation of both, respectively, as our baselines. Then, we concatenate angular features to each of these three to compare the performance. We evaluate the accuracy with two data streams, \ie{} angular static and velocity. 
We observe that all the feature types in both data streams receive accuracy boosting in response to incorporating angular features. For the static stream, concatenating angular features with the concatenation of joint and bone features leads to the most significant enhancement. As to the velocity stream, although the accuracy is lower than that of the static one, the improvement resulting from angular features is more substantial. In sum, concatenating all three features using the static data stream results in the highest accuracy.  

\textbf{Training Solely with Angular Encoding.}
We are interested in the performance of the network when only feeding the angular encoding, \ie{} no joint and bone features are used. The outcome is shown as the first row of \autoref{table:ang_feature_ens}, denoted as \emph{Ang}. We see training merely with angular encoding even outperforms that of utilizing the joint feature, indicating the completeness of angular encoding for depicting human skeleton motion trajectories. 

\textbf{Ensembling with Angular Encoding.}
We also study the change in accuracy when ensembling a network trained solely with angular features \emph{Ang} with networks trained with joint and bone features, respectively, as well as their ensemble. The results are reported in \autoref{table:ang_feature_ens}.
We obtain the accuracy of the above three models as the baseline results for each stream and compare them against the precision of ensembling the baseline models with \emph{Ang}.
We note that ensembling \emph{Ang} consistently leads to an increase in accuracy. As with the concatenation studies, angular features are more beneficial for the velocity stream.
However, unlike the case with concatenation, the accuracy of the two streams is similar. We also observe that ensembling with \emph{Bon} achieves considerable accuracy gain. An ensemble of \emph{Jnt}, \emph{Bon} and \emph{Ang} results in the highest accuracy in the static stream.

\textbf{Evaluating Angular Encoding of Each Category. }
We independently evaluate the boost of the angular encoding of the four categories, \ie local, center-oriented, pair-based, and finger-based. The utilized model is the BSL architecture. We discover that all these four categories can individually boost the recognition accuracy, as shown in \autoref{table:eval_each_ang}. Furthermore, the proposed angular encoding has been leveraged in an open challenge and revealed to be effective\footnote{In ICCV 2021, the winning team of a skeleton-based action recognition challenge leveraged the angular encoding proposed in this paper, achieving the 1st-place accuracy among 70+ teams. The utilized dataset was a newly collected skeleton dataset with drones. The winning team specifically evaluated the boost of accuracy from using our proposed angular encoding on the newly recorded dataset, showing the effectiveness of angular encoding. See their  \href{https://sutdapac-my.sharepoint.com/:v:/g/personal/jun_liu_sutd_edu_sg/EYL7JHl-f2FPvo95wmyoX88Bp_pn71wjgv6J4uiIZGACeA?e=UfLYPs}{presentation (clickable)} at 8:30. }.

\subsection{Comparison with State of the Art Models}
The ablation studies indicate fusing angular features in both concatenating and ensembling forms can boost accuracy. Hence, we include the results of both approaches as well as their combination in \autoref{tab:compare_with_sota}. In practice, the storage and the run time may become bottlenecks. Thus, we consider not only the recognition accuracy but also the number of parameters (in millions) and the inference time (in gigaFLOPs). 
The unavailable results are marked with a dash. 

We achieve new state-of-the-art accuracies for recognizing skeleton actions on both datasets, \ie{} NTU60 and NTU120. For NTU120, \ourmodel{} outperforms the existing state-of-the-art model by a wide margin. 

Apart from the higher accuracy, \ourmodel{} requires fewer parameters and a shorter inference time. We evaluate the inference time of processing a single NTU120 action video for all the methods. Compared with the existing most accurate model, \ourmodel{} requires fewer than 70\% of the parameters and less than 70\% of the run time while achieving \emph{higher} skeleton-based recognition results.  

Of note, the proposed angular features are compatible with the listed competing models. If one seeks even higher accuracy, the employed simple GCN can be replaced with a more sophisticated model, such as MS-G3D~\cite{liu2020disentangling}, although this change can lead to more parameters and longer inference time. For example, if we employ more complicated MSG3D~\cite{liu2020disentangling} instead of our MSGCN, the accuracy can be further improved as \autoref{tab:msg3d_compare} shows. Nonetheless, both the number of parameters and the GFlops will also correspondingly increase.

\section{Analysis of Angular Encoding}
We want to provide an intuitive understanding of how angular features help in differentiating actions. To this end, we compare the results from two models trained with the joint features and the concatenation of joint and angular features. 

\subsection{Utilizing of All Types of Angular Encoding}
First, we concatenate all kinds of angular encoding with joint features and train the baseline network. The results are illustrated in \autoref{table:confusion}.  
We observe two phenomena: (a) the majority of the action categories receiving a substantial accuracy boost from angular features are hand-related, such as making a victory sign vs thumbs up. We hypothesize that the enhancement may result from our explicit design of angles for hands and fingers, so that the gestures can be portrayed more comprehensively. (b) for some actions, after the angular features have been introduced, the most similar actions change. 
This suggests that the angles are providing complementary information to the coordinate-based representations. For the new actions that still confuse the network after using the angular encoding, they are also challenging for humans to differentiate them from their corresponding ground-truth actions by just observing skeletons. 
For better understanding, 
We provide some visual examples displaying the confusing actions whose mostly confused counterparts get altered after using angular encoding in \autoref{fig:visual_action_sequence}. Among them, folding paper and counting money are easily confused, and reading and writing are also likely to be mixed up. 
We see these confusing pairs of skeletons are visually similar to those of humans. 

\subsection{Contributions from Different Angle Types }

Next, we conduct ablation studies on different types of the proposed angular encoding for improving the accuracy of recognizing skeleton-based actions. The baseline accuracy is obtained merely using the joint feature. Then, we concatenate different types of angular encoding with the joint feature to evaluate the effectiveness of each encoding type. We study the effects of different types of angular features on improving the accuracy of recognizing actions. 

The results are depicted in \autoref{fig:citation_network_bar_chart}. We observe: 
i) the center-oriented angular encoding boosts the accuracy with the largest margin for both static and velocity input features; the increases are 1.01\% and 2.02\% respectively. Since the center-oriented encoding reflects the distance from the joint to the body center, the results imply knowing such a distance is greatly beneficial to recognizing skeleton-based actions. This is consistent with our daily experience. To illustrate, people normally pose the hand farther away from the body center for the victory sign than for the ok sign. 
ii) Angular encoding improves more accuracy for the velocity input features than the static joint coordinates. The average improvements are 0.58\% and 1.42\% respectively. This difference indicates angular encoding provides more additional information in capturing the dynamic motion trajectories of actions than depicting the spatial structural information. iii) The part-based angular encoding only marginally heightens the accuracy of using the static features, only 0.22\%, whereas the increase improves substantially enlarges to 1.47\% for the velocity input. We conjecture this is because the actions performed by arms and legs involve a lot of dynamics. Thus, when using the velocity input, angular encoding provides complementary dynamic information to these actions. 

We investigate how each kind of angular encoding improves accuracy. To this end, we collect the top seven actions whose accuracy is improved by the angular encoding the most. The results are exhibited in \autoref{table:diff_angle_analysis}. We see: i) Equipping the velocity features with angular encoding boosts substantial accuracy for the long-lasting actions, such as `staple book'. In contrast, for the static input, most actions whose accuracy is significantly improved are those that last for a short time, such as `thumb up'. ii) The majority of actions whose accuracy is improved by a type of angular encoding are those performed by the anchor joints corresponding to the angular encoding. To illustrate, the finger-based encoding increases accuracy for the hand-related actions, while the part-based encoding benefits the actions heavily using arms and legs.

\section{Generalisability of Angular Encoding}
A possible concern is the generalisability of the proposed angular encoding. That is, will fusing angular encoding improve the accuracy of other backbone architectures? To answer this, we conduct experiments fusing angular encoding with the joint feature and feed the concatenated input to three recently-proposed backbone networks: ShiftGCN~\cite{cheng2020skeleton}, DecoupleGCN~\cite{chengdecoupling} and MSG3D~\cite{liu2020disentangling}. The utilized dataset is the cross-subject setting of NTU120. 

We display the results in \autoref{fig:diff_backbone}. We not only demonstrate the accuracy of fusing all kinds of proposed angular encoding, but we also separately concatenate every type of encoding with the joint feature and report the corresponding accuracy. We see fusing angular encoding with the original features consistently improves the accuracy of all three backbones. On the other hand, the effectiveness of different angular encoding varies in boosting accuracy. We observe the center-oriented angular encoding increases accuracy with the largest magnitude. Furthermore, angular encoding improves accuracy more when deployed in the velocity domain than in the static domain. These two observations are consistent with those on our simple backbone network. For DecoupleGCN, the part- and finger-based angular encoding more substantially improve accuracy than they do for our simple backbone. Specifically, although feeding the velocity input to DecoupleGCN initially leads to lower accuracy than using the static feature, the situation is reversed after fusing with these two types of angular encoding. 
These scenarios imply that using features in the velocity domain surpasses using the static joints.

\section{Discussion}

As we have described in the introduction, current GCNs are designed to extract features between two adjacent nodes. On the other hand, the angular features are higher-order ones beyond two adjacent vertices. We can theoretically view every angle as a hyperedge $e(v_1, v_2, v_3)$, where $v_1$, $v_2$ and $v_3$ are the constitutional joints of an angle. The angular encoding is their associated feature. The angular encoding extends the capability of existing GNNs to capture features of hyperedges. 

From the perspective of treating a skeleton as a hypergraph, we have proposed four categories of hyperedges. In contrast, existing work that also makes use of angle features only contains one type of hyperedges. 
\section{Conclusion}

To extend the capacity of GCNs in extracting body structural information, we propose higher-order representations in the form of angular features, 
The proposed angular features comprehensively capture the relative motion between different body parts while maintaining robustness against variations of subjects. Hence, they are able to
discriminate between challenging actions having similar motion trajectories, which causes problems for existing models. Our experimental results show that the angular features are complementary to existing features, \ie{} the joint and bone representations. By incorporating our angular features into a simple action recognition GCN, we achieve new state-of-the-art accuracy on several benchmarks while maintaining lower computational cost, thus supporting real-time action recognition on edge devices.

% \newpage
% \clearpage
\bibliographystyle{named}
\bibliography{refs}

% \newpage
% \clearpage
% \input{biography}

% \newpage
% \appendix
% \input{secs/appendix}

\end{document}

% \maketitle
\begingroup
\let\clearpage\relax 
% \maketitle
\onecolumn 
\section{Definitions of the First- and Second-Order Features}

We extract three different orders of features from the skeleton. The first-order features, which represent the coordinate of each joint, are directly extracted from the skeleton. The second-order features, which represent the direction and length from one joint to an adjacent joint, are computed by calculating the difference between the joint coordinates. If a joint has more than a single adjacent node, we choose the joint closer to the body's center. For example, given an elbow joint, we use the vector from the elbow to the neck in preference to the vector from the elbow to the wrist. In formula, bone feature $b(v)$ for joint $v$ is defined as: 
\begin{align}
    b(v) = b(v') - b(v), 
\end{align}
where $v'$ is the selected adjacent joint of joint $v$. That is, the representation of bone $\vec{b}_{vv'}$ is: 
\begin{align}
    \vec{b}_{vv'} = (x_{v'} - x_v, y_{v'} - y_v, z_{v'} - z_v), 
\end{align}
where $x_k,y_k,z_k$ represent the coordinates of joint $k$. Eventually, for each action instance, we can construct a feature tensor $X \in \mathbb{R}^{C \times T \times V \times M}$, where $C$, $T$, $V$ and $M$ respectively correspond to the numbers of channels, frames, joints, and participants.

\endgroup

% \clearpage
% \bibliographystyle{named}
% \bibliography{ijcai21}